\def\year{2020}\relax
\documentclass[letterpaper]{article} 
\usepackage{aaai20}  
\usepackage{times}  
\usepackage{helvet} 
\usepackage{courier}  
\usepackage[hyphens]{url}  
\usepackage{graphicx} 
\usepackage{graphicx}  
\newcommand{\citet}[1]{\citeauthor{#1} \shortcite{#1}}
\title{An End-to-End Dialogue State Tracking System with\\ Machine Reading Comprehension and Wide \& Deep Classification}
\author{Yue Ma, Zengfeng Zeng, Dawei Zhu, Xuan Li\\
\Large \textbf{Yiying Yang, Xiaoyuan Yao, Kaijie Zhou, Jianping Shen}\\
Pingan Life Insurance of China, Ltd\\
\{mayue241, zengzengfeng277, zhudawei446, lixuan208\}@pingan.com.cn\\
\{yangyiying283, yaoxiaoyuan617, zhoukaijie002, shenjianping324\}@pingan.com.cn}

\begin{document}

\maketitle

\begin{abstract}
This paper describes our approach in DSTC 8 Track 4: Schema-Guided Dialogue State Tracking. The goal of this task is to predict the intents and slots in each user turn to complete the dialogue state tracking (DST) based on the information provided by the task's schema. Different from traditional stage-wise DST, we propose an end-to-end DST system to avoid error accumulation between the dialogue turns. The DST system consists of a machine reading comprehension (MRC) model for non-categorical slots and a Wide \& Deep model for categorical slots. As far as we know, this is the first time that MRC and Wide \& Deep model are applied to DST problem in a fully end-to-end way. Experimental results show that our framework achieves an excellent performance on the test dataset including 50\% zero-shot services with a joint goal accuracy of 0.8652 and a slot tagging F1-Score of 0.9835.
\end{abstract}

\section{Introduction}
The goal-driven dialogue system aims to provide a human-computer interaction based on natural language. It has a wide range of applications in intelligent customer services and intelligent personal assistants. A dialogue system is composed of three modules: natural language understanding (NLU), dialogue management (DM) and natural language generations (NLG) \cite{chen2017survey}. DM is the core component of goal-driven dialogue systems, which is responsible for tracking the dialogue states and controlling the conversation. An excellent DM can improve the user experience by optimizing the conversation flow and reducing the number of interactions. DM contains DST module and action selection module. In the traditional pipeline architecture for dialogue system, a dialogue state tracker usually estimates the state of conversation with hand-craft rules or neural networks, which is built on the NLU result of the current utterance and the result of previous DST \cite{jurafsky2000speech,schlangen2009general}. Different from traditional solutions, we propose an end-to-end approach to solve DST problems without using the information of previous DST.

\begin{figure*}[t]
\centering
\includegraphics[width=0.9\textwidth]{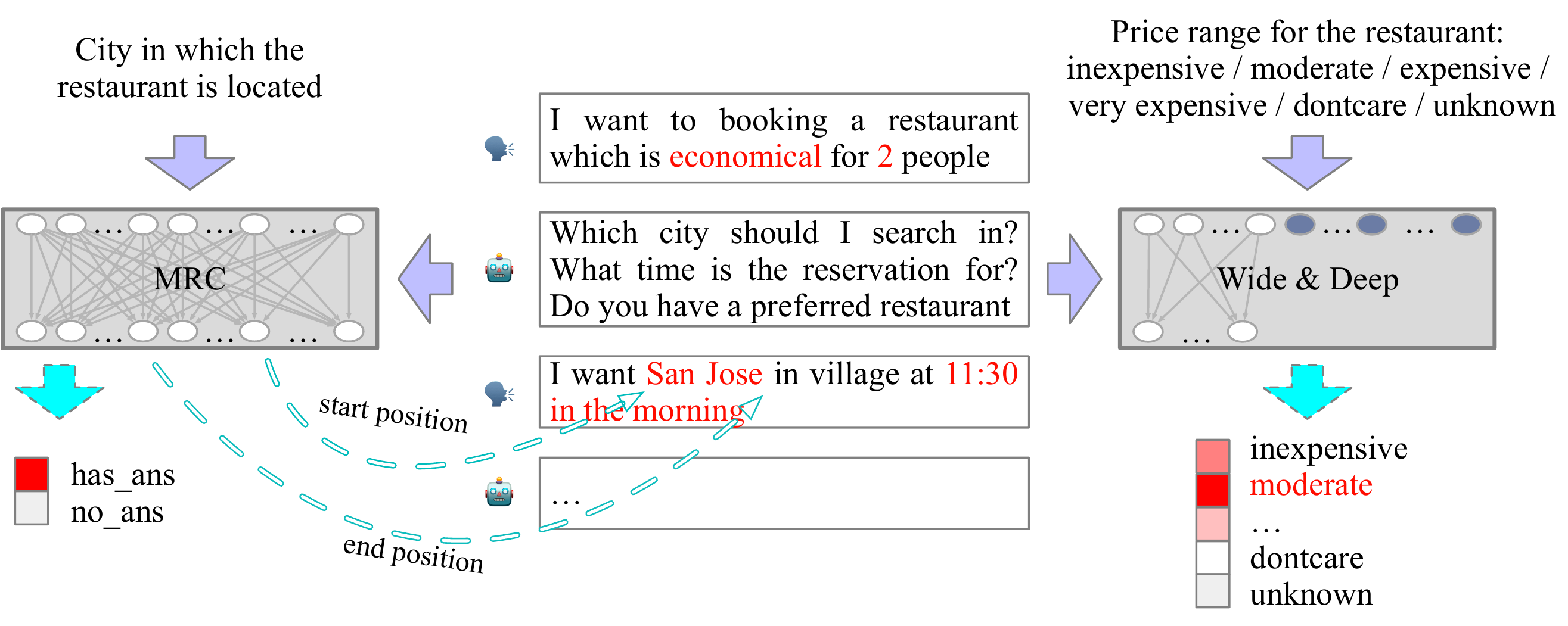} 
\caption{Dialogue sample and system framework}
\label{fig1}
\end{figure*}

In the multi-domain dialogue, users usually request different services in a conversation. In this situation, the dialogue system is required to seamlessly operate across all these services, which makes DST more challenging. Only when the dialogue state tracker accurately estimates the dialogue state across different domains, the dialogue system can make a right judgment for the subsequent action selection. The Schema-Guided Dialogue State Tracking track in the 8th Dialogue System Technology Challenge is proposed in this setting. In this competition, a series of services are pre-defined with the schema-guided approach. Different services are provided by different developers in each domain. The intents and slots of each service are accompanied by a sentence describing their semantics in natural language, where the semantics of the sentence can be recognized by the system and generalized in zero-shot learning to new schemas \cite{rastogi2019towards}. Figure \ref{fig1} shows a dialogue tracking example to attain such goal. Participants are required to track the state in each user turn. The dialogue state consists of these parts: the active intent, the requested slot, and the slot value. At the same time, the test dataset includes new services that do not exist in the training and the development dataset. The proposed system should be capable of tracking dialogue state for these new services.
The main challenges of DSTC 8 Track 4 are listed as follows:
\begin{quote}
\begin{enumerate}
    \item Unlike traditional DST systems, any string fragment of user or system utterance may become a slot value in this task, which makes the model unable to predict unseen slot value directly in a pre-defined value list. At the same time, a string fragment may be the answer to different slots in the multi-domain dialog system, which cannot be handled by the traditional DST systems with sequence labeling models.
    \item Since many services in the test dataset do not exist in the training dataset, the proposed model must have the ability to solve the problem of zero-shot service state tracking.
    \item In each user turn, one has to predict the intents and slots of the specific service. If the prediction is wrong at a certain turn, the error will be inherited by the subsequent state in the traditional DST system, and thus reduces the joint goal accuracy. 
\end{enumerate}
\end{quote}

To solve the above three problems, we propose two independent models to trace the dialogue state of all slots and intents at each user turn. The main contributions of our DST system include:
\begin{quote}
\begin{enumerate}
    \item We propose an end-to-end DST framework without inheriting information from previous DST results, which conquers the error accumulation issue of traditional DST systems. Our fully end-to-end DST framework consists of a machine reading comprehension (MRC-DST) module for non-categorical slot prediction and a Wide \& Deep DST model (WD-DST) combining the large-scale pre-trained model with hand-craft features for categorical slot prediction.
    \item Both models feed the natural language description of intents and slots as input to recognize the semantics of intents and slots, rather than just treating them as labels. This allows zero-shot generalization to a new service. The experiments show that these models have a good result for zero-shot service prediction.
    \item In the feature engineering of WD-DST, we make use of data augmentation to improve the model performance and the ability of zero-shot generalization.
\end{enumerate}
\end{quote}

\begin{figure*}[t]
\centering
\includegraphics[width=0.95\textwidth]{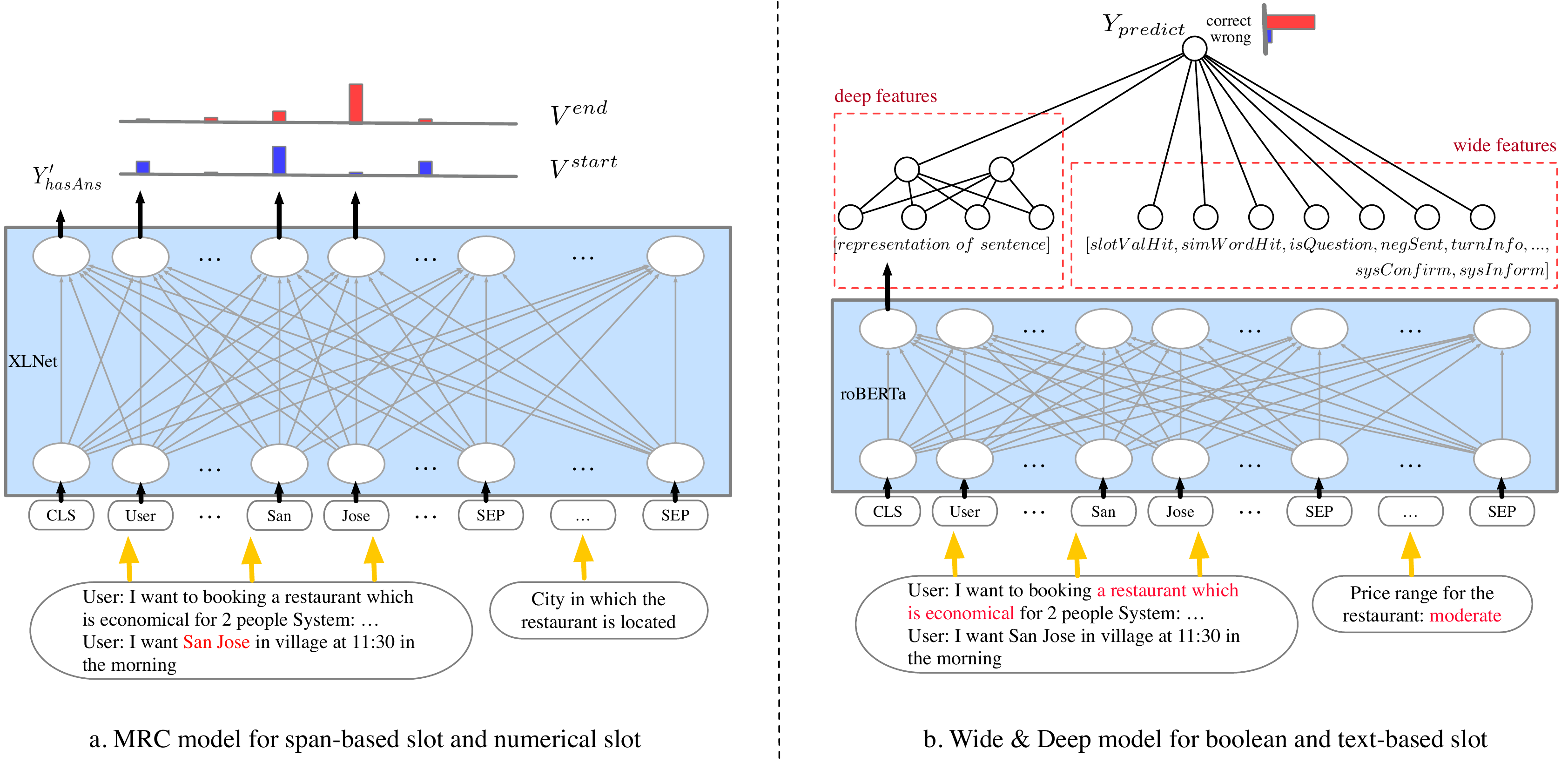} 
\caption{Structure of the MRC-DST and the WD-DST}
\label{fig2}
\end{figure*}

\section{Related Work}
Dialogue state tracking is an important module of dialogue management. It aims at describing the user's dialogue state at the current moment so that the system can select correct dialogue actions. The traditional DST system assumes that the candidate values of each slot are within a limit number. However, this assumption is not applicable to slots with unlimited number of values in advance. For zero-shot domains, it is more difficult to predefine the slot values in advance \cite{chen2017survey,mrksic-etal-2015-multi}.

Similar to the span-based slot in DSTC 8 Track 4, several researches are on converting fixed slot values into the substring of the dialogue context \cite{rastogi2017scalable,goel2018flexible,xu-hu-2018-end}. By this way, there is no need for value normalization for each slot. The traditional DST model uses a series of updating rules or several discriminant models, which are constructed according to the NLU results of the utterance and the previous DST results, to estimate the dialogue state at the current moment. Designing rules is cumbersome and the error accumulates as the dialogue continues. For this problem, researchers have proposed many neural networks to complete DST task in a reasonable way. \citet{perez2016dialog} use a memory network as a state tracker to implement DST module. As far as we know, this is the first study to treat the DST problem as a reading comprehension problem. However, the logic of the memory network is too complicated, and this method can only solve the problem in a fixed vocabulary set. \citet{gao2019dialog} use a simple attention-based RNN to point to slot values within the dialogue, but their system uses three different model components to make a sequence of predictions. Their stage-wise system has an error accumulation problem and cannot deal with the zero-shot problem effectively. \citet{wu-etal-2019-transferable} use TRADE-DST to track the slot value, their model is based on generate and copy mechanism and built upon bidirectional LSTM \cite{hochreiter1997long}. \citet{chao2019bert} use BERT-DST to solve the state tracking problem, but their model can only process one turn of dialogue in each time, while some additional dialog state update mechanism are required when completing the state tracking problem.

The 8th DSTC Track 4 competition began in July 2019. The test phase started on October 7th, and the evaluation of the game was closed on October 14th. Concurrently to our work, we notice that \citet{zhang2019find} release their research on multi-domain state tracking on October 8th. They use BERT as the encoder for dialogue history, use classification approach to solve list-based slot problems and use MRC approach to solve span-based slot problems. Their solution is similar to what we do in this competition. But our solution differs from theirs in several ways. First of all, we do not jointly train the models of these two tasks. Second, we convert the numerical slot into a span-based slot and solve it with an MRC model. Third, we use a Wide \& Deep classification model to solve boolean and text-based slot tracking problems. Fourth, we use the whole dialogue history as input to avoid error accumulation, while they use fixed length of the dialogue history as input.

The MRC model and the data augmentation technology are utilized in our proposed solution. MRC can be treated as an extension of the question-answering task, which means implementing a model to read an article and to answer questions related to it \cite{rajpurkar2016squad}. Since 2018, several large-scale pre-trained models based on Transformer \cite{vaswani2017attention} have been proposed. BERT \cite{devlin2018bert} and RoBERTa \cite{liu2019roberta} are based on Transformer, XLNet \cite{yang2019xlnet} is based on Transformer-XL. Recent researches demonstrate that Transformer-XL can model longer text dependency information \cite{yang2019xlnet}. We use XLNet as the base model for MRC to implement slot value extraction in our solution because of its outstanding performance in both SQuAD 2.0 \cite{rajpurkar2018know} and RACE \cite{lai2017race}. We use open-source translation and synonym services provided by Google, Baidu and Youdao, and use a back translation model between English and Chinese based on Transformer for data augmentation \cite{edunov-etal-2018-understanding}. Data augmentation enhances the model's ability to tackle the challenge of zero-shot problem, which is one of the biggest challenges for DSTC 8 Track 4. Experiment results show that our proposed model achieves a great improvement after using data augmentation.

\section{Model}
In Figure \ref{fig1}, in each user turn, we should predict the user slot state up to now. There are two kinds of slots for all services: categorical slots and non-categorical slots.

The non-categorical slot, which is also called as a span-based slot, as it is usually filled by the sub-string from the context. Thus this type of slots confines its starting and ending position in the context in the schema, we tackle the span-based slot tracking problem with a MRC model.

The categorical slot, which is also called as a list-based slot, is accompanied by a list consisting of all possible slot values in the schema. All predicted list-based slot values are within the pre-defined list. Through analysis, we found that all list-based slots can be divided into three types: boolean slots, text-based slots, and numerical slots. 1) Boolean slots consist of one value in \{True, False, dontcare, unknown\}. Apart from the candidate values provided in the schema, two additional candidate values, ``dontcare'' and ``unknown'', are added to meet the requirements of state tracking for each list-based slot, with additional key information presenting in the slot description, such as ``whether entrance to attraction is free''. 2) The key information of text-based slot comprises the slot description and the slot value, such as ``Name of airline'' and ``Delta Airlines''. 3) Numerical slots, different from the other two kind of slots, contain values directly in the dialog context. The value of a numerical slot is converted to Arabic numerals only when filling the answer. This finding for numerical slots is surprising given that this type of slot can be solved in a more reasonable way, just like span-based slots. Hence, in later experiments, we convert all numerical slots into the span-based slot format and treat them as non-categorical slots. For boolean and text-based slots, we use a Wide \& Deep model to process them.

Figure \ref{fig2} shows the architecture of the two models in our system. Figure \ref{fig2}(a) shows using the MRC-DST model to solve the slot tracking problems of span-based slot and the numerical slot. Figure \ref{fig2}(b) shows the WD-DST model for boolean slot and text-based slot tracking problem. The DST task is achieved by a combination of these two models.

\subsection{MRC-DST for Span-based Slots and Numerical Slot}
In Figure \ref{fig2}(a), we use XLNet proposed in \citet{yang2019xlnet} as the MRC model due to its ability to handle inputs of arbitrary length and its excellent performance in MRC tasks, such as SQuAD 2.0 and RACE. The model represents the input question and dialogue context as a single packed sequence. Suppose $S = [s_1, s_2, ..., s_n]$ is the dialogue context and $D = [d_1, d_2, ..., d_m]$ is the natural description of each slot. We concatenate them as the input of XLNet and get their corresponding representations $R$, where $[CLS]$ is a special token added in front of every sample, and $[SEP]$ is a special separator token. An answerable classifier is built upon the $[CLS]$ encoding and the start position encoding. A start vector $V^{start}$ and an end vector $V^{end}$ are introduced to predict the answer span during fine-tuning. We can formalize the process as:
\begin{equation}
    R = XLNet([CLS] \oplus S \oplus [SEP] \oplus D \oplus [SEP])
\end{equation}
\begin{equation}
    V^{start} = softmax(FC_{start}([r_1, r_2, ..., r_n]))
\end{equation}
\begin{equation}
    V^{end} = softmax(FC_{end}([r_1, r_2, ..., r_n] \oplus r_{startPos}))
\end{equation}
\begin{equation}
    Y'_{hasAns} = W_{hasAns} \cdot g([r_{startPos} \oplus r_{CLS}])
\end{equation}
\begin{equation}
    g(x) = tanh(W \cdot x + b)
\end{equation}
where $FC(\cdot)$ is a dense layer, $r_{startPos}$ is the representation of the first token in the answer span. We apply a logistic regression loss to predict the ability to answer question, together with a standard span extraction loss for question answering. We can calculate the loss as:
\begin{equation}
    \mathcal L_{hasAns} = \sum -log(Y'_{hasAns} \cdot Y_{hasAns})
\end{equation}
\begin{equation}
    \mathcal L_{start} = \sum -log(V^{start} \cdot Y_{start}^\top)
\end{equation}
\begin{equation}
    \mathcal L_{end} = \sum -log(V^{end} \cdot Y_{end}^\top)
\end{equation}
\begin{equation}
    \mathcal L_{total} = \mathcal L_{hasAns} + \mathcal L_{start} + \mathcal L_{end}
\end{equation}

\subsection{WD-DST for Boolean Slot and Text-based Slot}
In Figure \ref{fig2}(b), we use a Wide \& Deep model to solve the boolean and text-based slot tracking problems. Wide \& Deep model is used in recommendation systems \cite{cheng2016wide}, its advantage is to combine the features of two different sources and make prediction together. The model in Figure \ref{fig2}(b) is a joint model using deep features from RoBERTa and wide features from traditional feature engineering. RoBERTa \cite{liu2019roberta} is a deep matching feature extractor whose input comprises two parts. The first part is the same as XLNet, which is the history $S = [s_1, s_2, ..., s_n]$ of the dialogue up to the current moment. The second part $D'$ is the splicing of the natural language description of the slot and the candidate slot value to be judged, noted as ``description of slot: slot\_val''. These two parts serve as the input of RoBERTa together. The last layer representation of the $[CLS]$ is treated as the deep matching features.

Unlike span-based slots, which focus on time, location and name, boolean slot and text-based slot cover a wide variety of categories, including ``Type of cultural event'', ``Whether the flight is a direct one'', and ``The company that provides air transport services'' etc. . The variety of questions put higher demands on the transferring ability of the proposed model. To better improve the model performance, we use data augmentation to synonymize key vocabulary and slot values. The synonyms from external APIs and the top-10 back translation results of each slot value are added to its similar word set. For example, ``Theater"'', one of the ``event\_type'' slot values in ``Events\_3'', can be extended by using similar words and back translation and finally get about 67 synonymous words such as ``surgery'', ``stage'', ``cinema'', ``acting'', ``performing'', ``the stage'', ``drama'', ``the third acts'', ``show business'' and ``broadway''. With the help of these synonyms, we can better solve the zero-shot problem by constructing a number of linguistic features. Moreover, we introduce some discrete features of conversational state with the deep feature as the input of the topmost DNN to enhance the ability of our model. Some important and basic conversational features can be categorized as follows:
\begin{quote}
\begin{itemize}
    \item The tone of the utterance: interrogative, negative, declarative (by using regular expressions which generated from training dataset).
    \item Whether the utterance contains boolean slot descriptions or synonyms of these descriptions.
    \item Whether the utterance contains the values of text-based slots and synonyms of these slot values.
    \item Does the slot exist in the previous system actions? If it exists, which action? INFORM, REQUEST, OFFER or CONFIRM?
    \item The answer is yes or no.
    \item Whether the slot appears in requested slots in the current turn?
    \item The information of the current turn throughout the whole dialogue, e.g., turn number.
    \item Whether the slot is mentioned in the history?
\end{itemize}
\end{quote}

In total, 83 hand-crafted discrete features are employed as wide features. A full connected layer is used to transform the $[CLS]$ embedding. Finally, the 768-dimensional deep feature is stitched together with wide features as the uppermost layer input and the model is fine-tuned jointly. The process can be formalized as:
\begin{equation}
    R' = RoBERTa([CLS] \oplus S \oplus [SEP] \oplus D' \oplus [SEP])
\end{equation}
\begin{equation}
    I = DNN(r'_{CLS}) \oplus DiscreteFeatures
\end{equation}
\begin{equation}
    Y_{predict} = LogisticRegression(I)
\end{equation}
\begin{equation}
    \mathcal L_{w\&d} = CrossEntropy(Y_{predict}, Y_{true})
\end{equation}
We sort the classification scores of all candidate slot values, taking the value corresponding to the highest score as the final slot value.

\section{Experiment}

\begin{table*}[t]
\normalsize
\caption{Distribution of datasets}\smallskip
\centering
\begin{tabular}{l|l|l|l|l|l}
& Dialogues & Domains & Services & Zero-shot domains & Zero-shot services\\
\hline
Train & 16142 & 16 & 26 & - & -\\
Development & 2482 & 16 & 17 & 1 & 8\\
Test & 4201 & 18 & 21 & 3 & 11\\
\end{tabular}
\label{table1}
\end{table*}

\subsection{Data Sets and Indicators}
Table \ref{table1} lists the basic characteristics of the datasets. It is shown that about 50\% of the services are zero-shot services in the test dataset. This is a key challenge of the task.

Joint goal accuracy is the primary evaluation metric used for ranking submissions. Joint goal accuracy is the average accuracy of predicting all slot assignments for a user turn correctly. For non-categorical slots, a fuzzy matching score is used to reward partial matches with the ground truth. For categorical slot, a complete match between the ground truth and the predicted string declares success. The other evaluation indicators include active intent accuracy, slot tagging F1-score, requested slot F1-score and average goal accuracy. In this competition, contestants can use external datasets and resources, including using pre-trained models to improve the performance of the proposed system.

\subsection{Input Optimization}
We append a ``User:'' or a ``System:'' as tag before all utterances to help the model distinguish between system and user expressions, and then concatenate them in order as input. In the offline experiment, the result of adding the distinguishing tag is better than not adding the tag. In the latter experiments, all utterances in the dialogue history are added the distinguishing tag.

There are two main challenges during constructing our models: 1) unable to identify the slot value from a detected slot in the last sentence; 2) phone numbers tend to be the answer of numerical slots.

The first problem is that when we use XLNet for MRC, the slot value prediction results are usually wrong when the decisive information is at the end of the input, especially the last sentence of the whole interaction. For example, taking ``User: I want to booking a restaurant which is economical for 2 people System: Which city should I search in? What time is the reservation for? Do you have a preferred restaurant User: I Want San Jose in village at 11:30 in the morning'' as an input, the model can predict that there is a ``city'' slot in this dialogue history, but it can not identify ``San Jose'' as the slot value. The issue exists when applying in both XLNet and RoBERTa. We cannot investigate more due to the time constraint, but use a simple and effective method to handle it, i.e., copying the last sentence and appending it to the original input. This input trick does not change the meaning of the original input, but increases the best F1-score of the MRC by 0.5\%. Similarly, applying padding tokens can theoretically achieves the same result, but copying the last sentence can directly increase the probability of hitting the answer.

The second problem, which is also common, is that the phone number interfere the numerical slots, such as ``star\_rating'' slot and ``number\_of\_days'' slot, to make the right answer. According to our analysis, the phone number generally appears as the information in system ``INFORM'' actions and does not enter the dialogue state of the user. Therefore, in the pre-processing stage, all phone numbers are replaced with a ``phone'' tag, which greatly reduces prediction error.

\subsection{Intent Model and Requested Slot Model}
The user intent recognition and user requested slots recognition are not core tasks in this competition. So we do not put many efforts into these two tasks, but only use a two-class RoBERTa-based classification model to solve them. We use the previous nine utterances together with each intent (or slot) description as the input to make a classification. The model has a maximum input length of 512, a batch size of 16, a learning rate of 2e-5, and a embedding dropout of 0.1. We only use the training dataset to train these models and choose the best model on the development dataset as the final model during ten epoch training. The results on the development and test datasets are shown in Table \ref{table2}. In the test dataset, the intent classification model achieves an average intent accuracy of 0.9482, and the requested slot model achieves an F1-score of 0.9847.

\begin{table}[t]
\normalsize
\caption{Results of intent classification and requested slot model}\smallskip
\centering
\begin{tabular}{l|l|l}
 & Development & Test\\
\hline
Average Intent Accuracy & 0.9859 & 0.9482\\
Requested Slots F1-score & 0.9769 & 0.9847\\
\end{tabular}
\label{table2}
\end{table}

\begin{table}[t]
\normalsize
\caption{Slot tagging F1-score for span-based slots}\smallskip
\centering
\begin{tabular}{l|l|l}
 & Development & Test\\
\hline
only training dataset & 0.9889 & 0.9828\\
+ development dataset & - & 0.9835\\
\end{tabular}
\label{table3}
\end{table}

\subsection{Span-based Slot and Numerical Categorical Slot}
For span-based slots, we convert data into MRC data format by using the start and end positions of the slot value provided in the datasets. For the numerical slot, the position of the slot span is not directly provided in the dataset. In the data pre-processing stage, we first restore the position information of all numerical slots by using regular expressions, and then convert the information of numerical slot into MRC data format as span-based slots. For the numerical slot value, there are two kinds of patterns in the utterance, one is English expression, the other is Arabic numeral. The pattern of the numerical slot value is kept the same until the final slot filling stage. In the final slot filling stage, all patterns of numerical slot value will be converted into Arabic numerals.

We perform experiments on TPU v3 using the XLNet large model and using the same experimental parameters as \citet{yang2019xlnet}. Table \ref{table3} shows the MRC-DST experiment results of the span-based slot. On the development dataset, we get 0.9889 slot tagging F1-score. Because there are some services only appear in the development and test dataset, such as ``Alarm'', we use the development dataset as a supplement and achieve the best slot tagging F1-score of 0.9835 on the test dataset.

Apart from focusing on the overall slot tagging F1-score, we also care about the performance of the zero-shot problem because MRC-DST is used to solve more than 70\% of the slot tracking problems. In the final evaluation results of test dataset, the slot tagging F1-score of our model in zero-shot services is 0.9818, and the average accuracy of span-base slots is 0.9745, which show that our model has the ability to transfer on the zero-shot problem. The model makes predictions by understanding the dialogue context and natural language descriptions of slots. Therefore, for the zero-shot service and even the zero-shot domain, our model has an excellent performance.

\begin{table}[t]
\normalsize
\caption{Uses different models to solve the effect of numerical slot in development set}\smallskip
\centering
\begin{tabular}{l|l}
 & Accuracy\\
\hline
RoBERTa & 0.9378\\
WD-DST & 0.9551\\
MRC-DST & 0.9858\\
\end{tabular}
\label{table4}
\end{table}

We also use three methods to solve the tracking problem of numerical slot, their performance on the development dataset are shown in Table \ref{table4}. The experiment results on the development dataset show that MRC is superior to other methods. Using the MRC model to solve the problem of the numerical slot has achieved the best results, the accuracy reachs 0.9858, more than 0.03 improvement over the Wide \& Deep model and compared to a 0.05 increase over the RoBERTa classification model.

\subsection{Boolean and Text-based Categorical Slot}
We use RoBERTa(-base) as the deep feature extractor and optimize the Wide \& Deep model using the Adam optimizer \cite{KingmaB14} with $\beta_1=0.9$, $\beta_2=0.9999$, $\epsilon=1e-6$ and $L2$ weight decay of 0.01. The batch size is 16, the learning rate is 2e-5, the word embedding uses 0.1 dropout and fp16 training is used. 

\begin{table}[t]
\normalsize
\caption{Results of WD-DST in development dataset}
\centering
\begin{tabular}{l|l|l|l}
Accuracy & Average & Boolean Slot & Text-based Slot\\
\hline
RoBERTa & 0.9335 & 0.9183 & 0.9411\\
WD-DST & 0.9791 & 0.9624 & 0.9874\\
\end{tabular}
\label{table5}
\end{table}

The results of WD-DST are shown in Table \ref{table5}. We compare the results between the Wide \& Deep model and the RoBERTa-based classification model to idenfy the effect of hand-crafted features. The RoBERTa-based classification model achieves an average accuracy of 0.9355 on the development dataset, while the Wide \& Deep model achieves 0.9751. Contrast with the single deep learning model, Wide \& Deep model can enhance the system performance greatly. This is because the input of the Wide \& Deep model adds a lot of hand-crafted semantic features, which is more effective than using RoBERTa model only. The combination of deep semantic understanding features and shallow linguistic features is the most important factor for boolean and text-based slot tracking tasks. Moreover, there is a problem with models that only use RoBERTa, if the dialogue history is too long, it will truncate some key information which resulting in a decrease of the model's accuracy.

\subsection{Overall Results}
The dialogue state usually can be shared within a service, but there exists a special case. When the slot value has conflict between different intents of one service, the dialogue state cannot be shared when intents are switched. For example, in the ``Payment\_1'' service, when two intents ``MakePayment'' and ``RequestPayment'' are switched, the dialogue state cannot be shared because the value of the ``amount'' slot is different for two intents. Based on our observations, we apply a special rule to the dialogue state tracker. Once we detect the new intent is turned on in ``Payment\_1'', we clear the history state information to avoid conflicts.

By using the above models and the special rule, the system we proposed achieves the result with joint goal accuracy of 0.8652 and slot tagging F1-score of 0.9835 on the test dataset.

\begin{table}[t]
\normalsize
\caption{The performance of our proposed system on the test dataset}
\centering
\begin{tabular}{p{5.5cm}|p{1.5cm}}
 & Joint Goal Accuracy\\
\hline
MRC-DST (span-based slot) + RoBERTa (list-based slot) & 0.7048\\
\hline
MRC-DST (span-based and numerical slot) + WD-DST (boolean and text-based slot) & 0.8052\\
\hline
MRC-DST (span-based and numerical slot) + WD-DST (boolean and text-based slot) + data augmentation & 0.8653\\
\end{tabular}
\label{table6}
\end{table}

We have done some limited ablation studies on the test dataset. The results are shown in Table \ref{table6}. The ensemble system using MRC-DST and WD-DST incorporating text augmentation achieves the best results, with more than 10\% improvement over other methods. This is because feature engineering is not limited by the length of dialogue history and many language features can be constructed artificially. In the experiment, we find that the MRC-DST has a strong semantic understanding ability. We also use deep learning semantic representation to increase the generalization ability for WD-DST. The optimal result can be achieved by combining the data augmentation technology. The experiment results show that data augmentation can improve the performance in the zero-shot services, which is consistent with our observation in the development dataset.

\begin{table}[t]
\normalsize
\caption{Performance of the Top 5 model on the test dataset}
\centering
\begin{tabular}{l|p{1.5cm}|p{1.8cm}}
Team Name & Joint Goal Accuracy & Slot Tagging F1-score\\
\hline
Team 9 (ours) & 0.8653 & 0.9835\\
Team 14 & 0.7726 & 0.9812\\
Team 12 & 0.7375 & 0.9763\\
Team 8 & 0.7344 & 0.7105\\
Team 5 & 0.7303 & 0.9692\\
\end{tabular}
\label{table7}
\end{table}

With these two end-to-end models, we achieve Top 1 ranking in the DSTC 8 Track 4 competition, and the final results are shown in Table \ref{table7}.

\section{Conclusion and Future Work}
In this paper, we describe our solution in the DSTC 8 Track 4 competition. We apply two RoBERTa classification model to solve intent recognition and requested slot recognition. We propose an end-to-end MRC-DST model to solve the state tracking problem of span-based slot and numerical slot. MRC-DST uses the entire dialogue history as input and takes the slot description as a question to extract the answer. Through the complete end-to-end approach, we achieve a slot tagging F1-score of 0.9835, which is the highest of the participating teams. For boolean slot and text-based slot, we use a WD-DST model to solve it. All the intent and slot name are replaced by their natural language description as input, so that our proposed system can solve the zero-shot problem.

In the future, we are going to put more efforts on exploring a unified model, e.g., a unified reading comprehension model, to track the states of all slots. The pre-trained models tend to make wrong predictions when the key information exists at the end of the input. We ease this issue by the copying and appending mechanism. We plan to investigate more on it and develop models to improve the interpretability of pre-trained models.

\bibliographystyle{aaai}
\bibliography{dstc8track4mrcdst}
\end{document}